\documentclass[letterpaper, 10 pt, journal, twoside]{ieeetran}
\usepackage{graphics}    %
\usepackage{times}       %
\usepackage{amsmath}     %
\usepackage{amssymb}     %
\usepackage{graphicx}
\usepackage{algorithm}
\usepackage[noend]{algpseudocode}
\usepackage{booktabs}
\usepackage{color} %
\usepackage{subfigure}
\usepackage{tikz}

\def\secref#1{Sec.~\ref{#1}}
\def\figref#1{Fig.~\ref{#1}}
\def\tabref#1{Tab.~\ref{#1}}
\def\eqref#1{Eq.~(\ref{#1})}
\def\algref#1{Alg.~\ref{#1}}

\makeatletter

\usepackage{xspace}
\DeclareRobustCommand\onedot{\futurelet\@let@token\@onedot}
\def\@onedot{\ifx\@let@token.\else.\null\fi\xspace}

\def\ie{i.e\onedot}

\def\etal{\emph{et al}\onedot}
\makeatother

\usepackage{array}
\newcolumntype{L}[1]{>{\raggedright\let\newline\\\arraybackslash\hspace{0pt}}m{#1}}
\newcolumntype{C}[1]{>{\centering\let\newline\\\arraybackslash\hspace{0pt}}m{#1}}
\newcolumntype{R}[1]{>{\raggedleft\let\newline\\\arraybackslash\hspace{0pt}}m{#1}}

\def\argmin{\mathop{\rm argmin}}

\newcommand{\norm}[1]{\|#1\|}

\def\Prob{P}

\title{Adaptive Robust Kernels for Non-Linear Least Squares Problems}
\author{Nived Chebrolu, Thomas L{\"a}be, Olga Vysotska, Jens Behley, and Cyrill Stachniss%
  \thanks{This work has partially been funded by the Deutsche Forschungsgemeinschaft (DFG, German Research Foundation) under Germany's Excellence Strategy, EXC-2070 - 390732324 - PhenoRob as well as by the EC under grant agreement no. 101017008 - Harmony.}
  \thanks{All authors are with the University of Bonn, Germany.}
}

\begin{document}
\maketitle

\begin{tikzpicture}[overlay, remember picture]
\path (current page.north east) ++(-4.2,-0.2) node[below left] { 
This paper has been accepted for publication in the IEEE Robotics and Automation Letters.
};
\end{tikzpicture}
\begin{tikzpicture}[overlay, remember picture]
\path (current page.north east) ++(-3.0,-0.6) node[below left] {
Please cite the paper as: Nived Chebrolu, Thomas L{\"a}be, Olga Vysotska, Jens Behley, and Cyrill Stachniss
};
\end{tikzpicture}
\begin{tikzpicture}[overlay, remember picture]
\path (current page.north east) ++(-6.0,-1) node[below left] {
``Adaptive Robust Kernels for Non-Linear Least Squares Problems'',
};
\end{tikzpicture}
\begin{tikzpicture}[overlay, remember picture]
\path (current page.north east) ++(-7.0,-1.4) node[below left] {
 \emph{IEEE Robotics and Automation Letters (RA-L)}, 2021.
};
\end{tikzpicture}

\begin{abstract}
State estimation is a key ingredient in most robotic systems. Often, state estimation is performed using some form of least squares minimization. Basically, all error minimization procedures that work on real-world data use robust kernels as the standard way for dealing with outliers in the data.
These kernels, however, are often hand-picked, sometimes in different combinations, and their parameters need to be tuned manually for a particular problem.  
In this paper, we propose the use of a generalized robust kernel family, which is
automatically tuned based on the distribution of the residuals and includes the common m-estimators. 
We tested our adaptive kernel with two popular estimation problems in robotics, namely
ICP and bundle adjustment. The experiments presented in this paper suggest that our approach provides higher robustness while avoiding a manual tuning of the kernel parameters.
\end{abstract}

\section{Introduction}
\label{sec:intro}

\IEEEPARstart{S}{tate} estimation is a central building block in robotics and is used in a variety of different components, such as simultaneous localization and mapping~(SLAM)~\cite{stachniss2016handbook-slamchapter}.  A large number of state estimation solvers perform some form of non-linear least squares minimization. Prominent examples are the optimization of SLAM graphs, the ICP algorithm, visual odometry, or bundle adjustment~(BA), which all seek to find the minimum of some error function. As soon as real-world data is involved, outliers will occur in the data. A common source of such outliers stems from data association mistakes, for example, when matching features.

To avoid that even a few such outliers have strong effects on the final solution, robust kernel functions are used to down-weight the effect of gross errors. Several robust kernels have been developed to deal with outliers arising in different situations. Prominent examples include the Huber, Cauchy, Geman-McClure, or Welsch functions that can be used to obtain a robustified estimator~\cite{zhang1997ivc}.

However, the proper choice of the best kernel for a given problem is not straightforward. As the robust kernels define the distribution from which the outliers are generated, their choice is problem-specific. In practice,  the choice of the kernel is often done in a trial and error manner, as in most situations there is no prior knowledge of the outlier process.  For some approaches such as bundle adjustment, today's implementations even vary the kernel between iterations or pair them with outlier rejection heuristics. Moreover, for several robotics applications such as SLAM, the outlier distribution itself changes continuously depending on the structure of the environment, dynamic objects in the scene and other environmental factors like lighting.  This often means that a fixed robust kernel chosen a-priori cannot deal effectively with all situations.

In this paper, we aim at circumventing the trial and error process for choosing a kernel and at exploring the automatic adaptation of kernels to the outliers online. To achieve this, we use a family of robust loss functions proposed by Barron~\cite{barron2019cvpr}, which generalizes several popular robust kernels such as  Huber, Cauchy, Geman-McClure, Welsch, etc.  The key idea is to dynamically tune this generalized loss function automatically based on the current residual distribution so that one can blend between such robust kernels and make the choice a part of the optimization problems.

\begin{figure}[t]
  \centering
  \includegraphics[width=\linewidth]{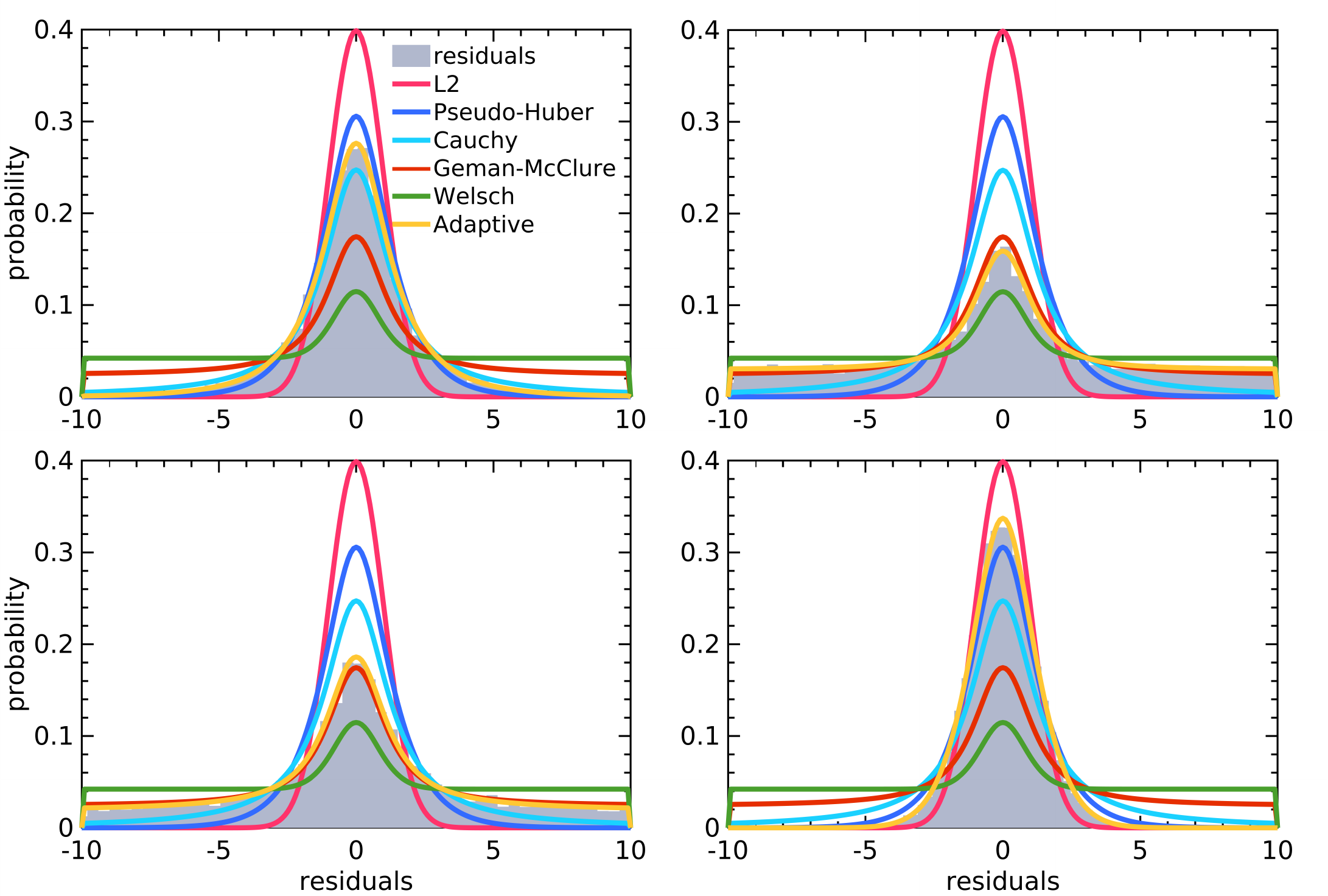}
  \caption{Probability densities of different robust kernels. The adaptive robust kernel (in yellow) is able to describe the actual residual distribution in different situations better than a fixed robust kernel for all cases. As a result, it provides better robustness to different types of outliers during the state estimation process.}
  \label{fig:motivation}
\end{figure}

The main contribution of this paper is an easy-to-implement approach for dynamically adapting the robust kernels in non-linear least squares (NLS) solvers, which builds on top of the generalized formulation of Barron~\cite{barron2019cvpr}. We achieve this by estimating a hyper-parameter for a generalized loss function, which controls the shape of the robust kernel.
This parameter becomes part of the estimation process and we determine it along with the unknown parameters of the model. We extend the usable range of this parameter compared to the formulation of Barron~\cite{barron2019cvpr}. This allows us to better deal with a larger set of outlier distributions compared to fixed kernels and to the Barron formulation. See \figref{fig:motivation} for a visualization.

In sum, we make the following key claims. Our approach can
(i)~perform robust estimation without committing to a fixed kernel beforehand, (ii)~adapt the shape of the kernel to the actual outlier distribution, and (iii)~illustrate the performance on two common example problems, namely ICP and bundle adjustment.

\section{Related Work}
\label{sec:related}

Robust kernels are the de-facto solution to perform state estimation using least-squares minimization in the presence of outliers. 
To deal with different outlier distributions, several robust kernels such as Huber, Cauchy, Geman-McClure, or Welsch have been proposed in the literature. 
Zhang~\cite{zhang1997ivc} and Bosse~\cite{bosse2016fntr} apply these kernels to different kind of estimation problems in vision and robotics.
Black and Rangarajan~\cite{black1996ijcv} investigate equivalence between robust loss minimization and outlier processes, and apply this idea to several vision problems such as surface reconstruction, segmentation, optical flow etc. 
Babin~\etal~\cite{babin2019icra} analyzed several popular robust kernels for registration problems and provide advice for using different kernels depending on the scenario.
Similar analysis and recommendations exists for visual odometry and BA in~\cite{mactavish2015crv},~\cite{zach2014eccv}. 
 
In this work, instead of choosing a specific robust kernel for a particular scenario, we dynamically adapt a robust kernel to the actual outlier distribution during the optimization process. To do this, we build upon the generalized kernel formulation recently proposed by Barron~\cite{barron2019cvpr} for training neural networks. It generalizes over popular robust kernels and we formulate an approximation of it for the use in NLS estimation. 

For pose graph SLAM problems, several approaches exist to deal with the outliers dynamically~\cite{stachniss2016handbook-slamchapter,agarwal2014ram}. S{\"u}nderhauf and Protzel~\cite{sunderhauf2012iros} propose introducing additional switch variables
to the original optimization problem, which determines whether an observation should be used or discarded during optimization. 
The RRR approach~\cite{latif2012rss} is a robust SLAM back-end that targets to reject false constraints through constraint clustering and mutually consistency checks.
Agarwal \etal~\cite{agarwal2013icra} propose a robust kernel, which dynamically weighs the observations without requiring to estimate any additional variables. 
Lajoie \etal~\cite{lajoie2019ral} and Yang \etal~\cite{yang2020arxiv} further this idea as a truncated least squares problem which can be solved efficiently as a semi-definite program. They also provide solutions with certain robustness guarantees for the registration and SLAM problem. Recently, Yang \etal~\cite{yang2020ral} propose a robust estimation framework based on graduated non-convexity (GNC) methods which solves a sequence of minimization problems which are convex initially, and converge eventually to the original non-convex robust loss.
 
Taking a probabilistic view, several robust kernels are understood to arise from a probability distribution, which can be used to determine the best kernel type based on the actual observations. Agamennoni \etal~\cite{agamennoni2015icra} propose to use an elliptical distribution to represent several popular robust kernels. They estimate hyper-parameters for each kernel type based on the residual distribution and perform a model comparison to determine the best kernel for the situation at hand. In this paper, we take a different approach and adapt the robust kernel shape by using the probability distribution of a generalized loss function~\cite{barron2019cvpr}. We do not require an explicit model comparison to choose the best kernel and estimate the kernel shape through an alternating minimization procedure.

\section{Least Squares with an Adapting Kernel}
\label{sec:main}

Our approach targets dynamically adapting robust kernels when solving NLS  problems by estimating a hyper-parameter that controls the shape of the robust kernel. This parameter becomes part of the estimation process in an alternating error minimization procedure. Before explaining our approach, we first explain robust NLS estimation and generalized kernels to give the reader a complete view~(\secref{sec:robust_estimation}). We then present the generalized robust kernel proposed by Barron~\cite{barron2019cvpr}, which is the foundation of our work~(\secref{sec:adaptive_kernel}). We extend Barron's robust kernel to deal with strong outliers typically encountered in robotics applications~(\secref{sec:truncated_kernel}) and use it for solving typical state estimation problems~(\secref{sec:optimization_alpha}). 

\subsection{Robust Least Squares Estimation}
\label{sec:robust_estimation}
Several state estimation problems in robotics involve estimating unknown parameters~$\theta$ of a model given noisy observations~$z_i$ with~$i = 1, \dots, N$. These problems are often framed as non-linear least squares optimization, which aims to minimize the squared loss: 
\begin{eqnarray}
  \label{eq:squared_loss}
  \theta^{*} &=& \argmin_{\theta} \frac{1}{2}\sum_{i=1}^{N} w_i \norm{r_i(\theta)}^2, 
\end{eqnarray}
where~$r_i(\theta)=f_i(\theta) - z_i$ is the residual and~$w_i$ is the weight for the~$i^\mathrm{th}$ observation. The estimate~$\theta^{*}$ is statistically optimal if the error on the observations~$z_i$ is Gaussian. In case of non-Gaussian noise, however, the estimate~$\theta^{*}$ can be arbitrarily poor~\cite{huber1964kernel}. To reduce this impact of outliers,  sub-quadratic losses can be applied. The main idea of a robust loss is to downweight large residuals that are assumed to be caused from outliers such that their influence on the solution  is reduced. This is achieved by optimizing: 
\begin{eqnarray}
  \label{eq:robust_loss}
  \theta^{*} = \argmin_{\theta} \sum_{i=1}^{N} \rho({r_i(\theta)}), 
\end{eqnarray}
where~$\rho(r)$ is also called the robust loss or kernel. Several robust kernels have been proposed to deal with different kinds of outliers such as Huber, Cauchy, and others~\cite{zhang1997ivc}. A summary of several popular robust kernels can be found in the work by MacTavish \etal~\cite{mactavish2015crv}.

The optimization problem in~\eqref{eq:robust_loss} can be solved using the iteratively reweighted least squares (IRLS) approach~\cite{zhang1997ivc}, which solves a sequence of weighted least squares problems. We can see the relation between the least squares optimization in~\eqref{eq:squared_loss} and robust loss optimization in~\eqref{eq:robust_loss} by comparing the respective gradients which go to zero at the optimum (illustrated only for the $i^\mathrm{th}$ residual):
\begin{eqnarray}
  \label{eq:gradient_robust}
  \frac{1}{2}\frac{\partial (w_i r^2_i(\theta))}{\partial\theta} 
  &=&
  w_i r_i(\theta) \frac{\partial r_i(\theta)}{\partial\theta} \\
  \label{eq:gradient_ls}
  \frac{\partial(\rho(r_i(\theta)))}{\partial\theta} 
  &=& 
  \rho'(r_i(\theta)) \frac{\partial r_i(\theta)}{\partial\theta}.
\end{eqnarray}

By setting the weight $w_i= \frac{1}{r_i(\theta)}\rho'(r_i(\theta))$, we can solve the robust loss optimization problem 
by using the existing techniques for weighted least-squares. This scheme allows standard solvers using Gauss-Newton and Levenberg-Marquardt algorithms to optimize for robust losses and is implemented in popular optimization frameworks such as Ceres~\cite{agarwal2010ceres}, g2o~\cite{kummerle2011icra}, and iSAM~\cite{kaess2008tro}. 

\subsection{Adaptive Robust Kernel}
\label{sec:adaptive_kernel}
\begin{figure}
  \centering
  \includegraphics[width=\linewidth]{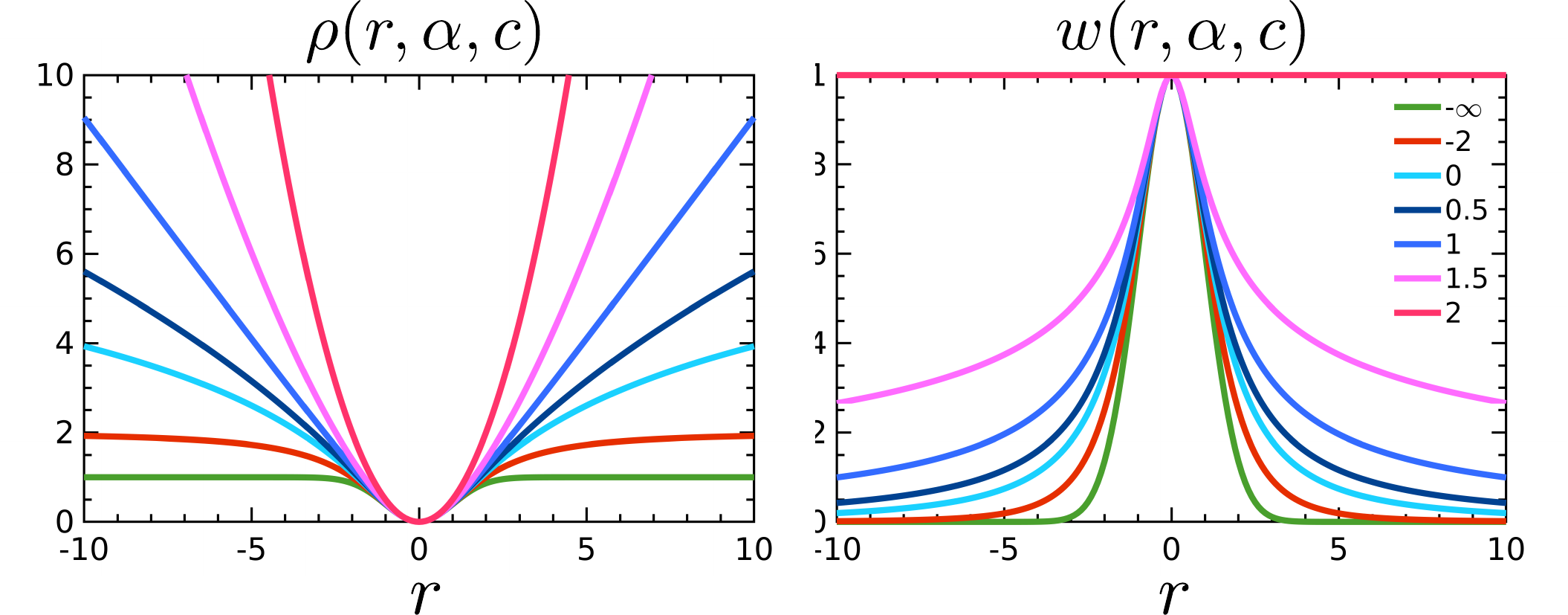}
  \caption{Left: General robust loss $\rho(r, \alpha, c)$  takes different shapes depending on the value of $\alpha$. 
  Right: Corresponding weights for kernels with different $\alpha$ values. A smaller $\alpha$ corresponds to a larger down-weighting of the residuals.}
  \label{fig:barron_loss}
\end{figure}

Barron~\cite{barron2019cvpr} proposes a single robust kernel that generalizes for several popular kernels such as pseudo-Huber/L1-L2, Cauchy, Geman-McClure, Welsch.
The generalized kernel~$\rho$ is given by:
\begin{eqnarray}
  \label{eq:barron_loss}
  \rho(r, \alpha, c)&=&\frac{|\alpha-2|}{\alpha}\left(\left(\frac{(r / c)^{2}}{|\alpha-2|}+1\right)^{\alpha / 2}-1\right),
\end{eqnarray}
where $\alpha$ is a real-valued parameter that controls the shape of the kernel and $c>0$ is the scale parameter that determines the size of quadratic loss region around $r=0$. Adjusting the parameter~$\alpha$  essentially allows us to realize different robust kernels. Some special cases are squared/L2 loss ($\alpha=2$), pseudo-Huber/L1-L2 ($\alpha=1$), Cauchy ($\alpha=0$), Geman-McClure ($\alpha=-2$), and Welsch ($\alpha=-\infty$). Note that $\rho(r, \alpha, c)$ in \eqref{eq:barron_loss} is not defined for $\alpha = 0$ and $\alpha =2$, and are instead defined as the respective pointwise limits at these values~(see Eq.~(8) of Barron's paper~\cite{barron2019cvpr}). 

The general loss function~$\rho(r, \alpha, c)$ and the corresponding weights curve~$w(r, \alpha, c)$ are illustrated in~\figref{fig:barron_loss} for several values of $\alpha$. The shape of the weights curve provides an insight into the influence that a residual has on the solution while minimizing the robust loss function in~\eqref{eq:robust_loss}. For example, for~$\alpha = 2$, the weights for all residuals are one, meaning that all residuals are treated the same. Whereas for~$\alpha = -\infty$, all residuals greater than $3c$ will not affect the solution $\theta^*$ significantly as they are weighed down to very small values.    

\begin{figure}[t]
  \centering
  \includegraphics[width=\linewidth]{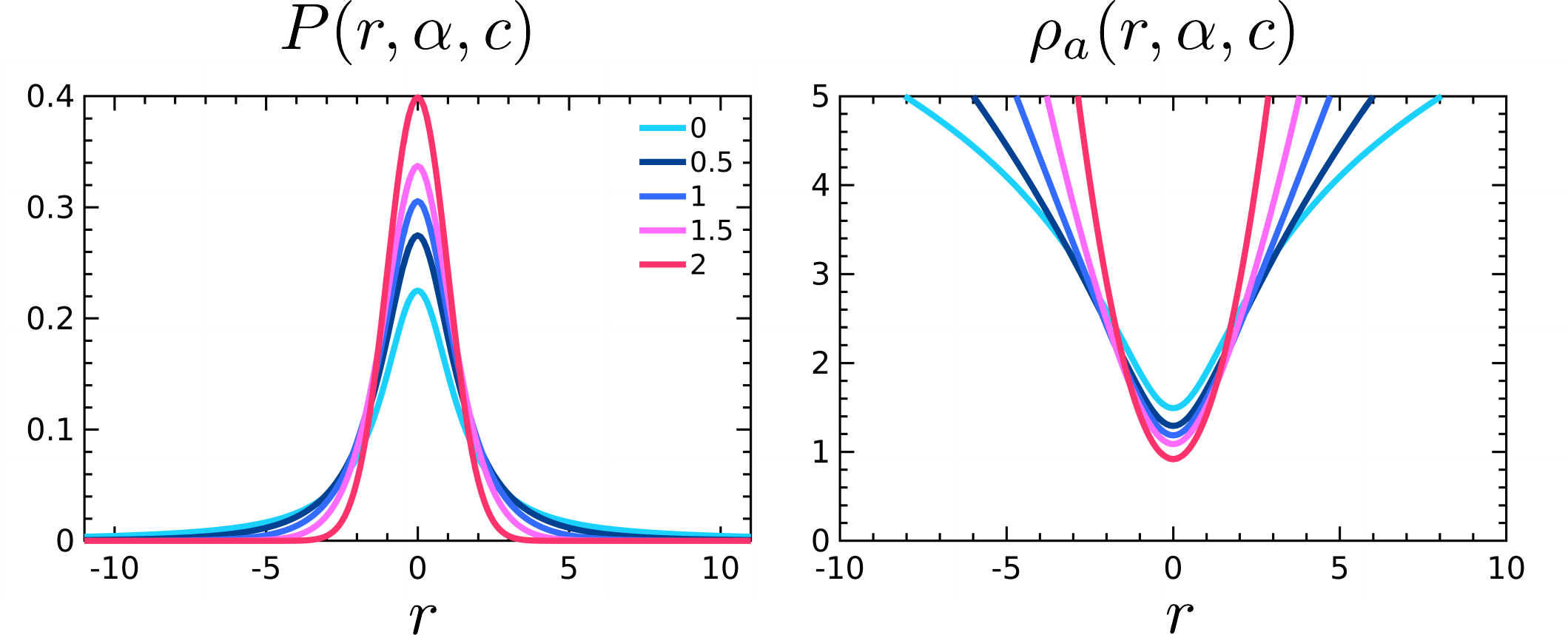}
  \caption{Left: Probability distribution $\Prob(r, \alpha, c)$ of generalized robust loss function for different values of $\alpha$. Right: Adaptive robust loss $\rho_a(r, \alpha, c)$ obtained as the negative log-likelihood of $\Prob(r, \alpha, c)$. This adaptive loss enables automatic tuning of $\alpha \in [0, 2]$. }
  \label{fig:adaptive_loss}
\end{figure}

With this generalized robust loss, we can interpolate between a range of robust kernels simply by tuning $\alpha$. To automatically determine the best kernel shape through the parameter~$\alpha$, we treat~$\alpha$ as an additional unknown parameter while minimizing the generalized loss: 
\begin{eqnarray}
\label{eq:joint_robust_loss}
  (\theta^{*},\alpha^{*})  &=& \argmin_{(\theta,\alpha)} \sum_{i=1}^{N} \rho({r_i(\theta)}, \alpha).
\end{eqnarray}

However, this optimization problem in \eqref{eq:joint_robust_loss} can be trivially minimized by choosing an~$\alpha$ that weighs down all residuals to small values without affecting the model parameters $\theta$, essentially treating all data points as outliers. Barron~\cite{barron2019cvpr} avoids this situation by constructing a probability distribution based on the generalized loss function~$\rho(r, \alpha, c)$ as
\begin{eqnarray}
\label{eq:barron_loss_pdf}{}
\Prob(r ,\alpha, c)&=&\frac{1}{c Z(\alpha)} e^{-\rho(r, \alpha, c)} \\
\label{eq:partition}
Z(\alpha)&=&\int_{-\infty}^{\infty} e^{-\rho(r, \alpha, 1)}\ dr,
\end{eqnarray}
where $Z(\alpha)$ is a normalization term, also called partition function, which defines an adaptive general loss as the negative log-likelihood of~\eqref{eq:barron_loss_pdf},
\begin{eqnarray}
\label{eq:adaptive_general_loss}
\rho_a(r,\alpha,c) &=& -\text{log}\,\Prob(r,\alpha, c) \\
                  &=& \rho(r,\alpha,c) + \text{log}\,cZ(\alpha).
\end{eqnarray}

The adaptive loss $\rho_a(\cdot)$ is simply the general loss $\rho(\cdot)$ shifted by the log partition. This shift introduces an interesting trade-off. A lower cost for increasing the set of outliers comes with a penalty for the inliers and vice versa. This trade-off forces the optimization in~\eqref{eq:joint_robust_loss} to choose a suitable value for $\alpha$ instead of trivially ignoring all residuals by turning every data point into an outlier. The probability distribution~$\Prob(r, \alpha, c)$ and the adaptive loss function are plotted in~\figref{fig:adaptive_loss} for visualization.

\subsection{Truncated Robust Kernel}
\label{sec:truncated_kernel}
\begin{figure}
  \centering
  \includegraphics[width=\linewidth]{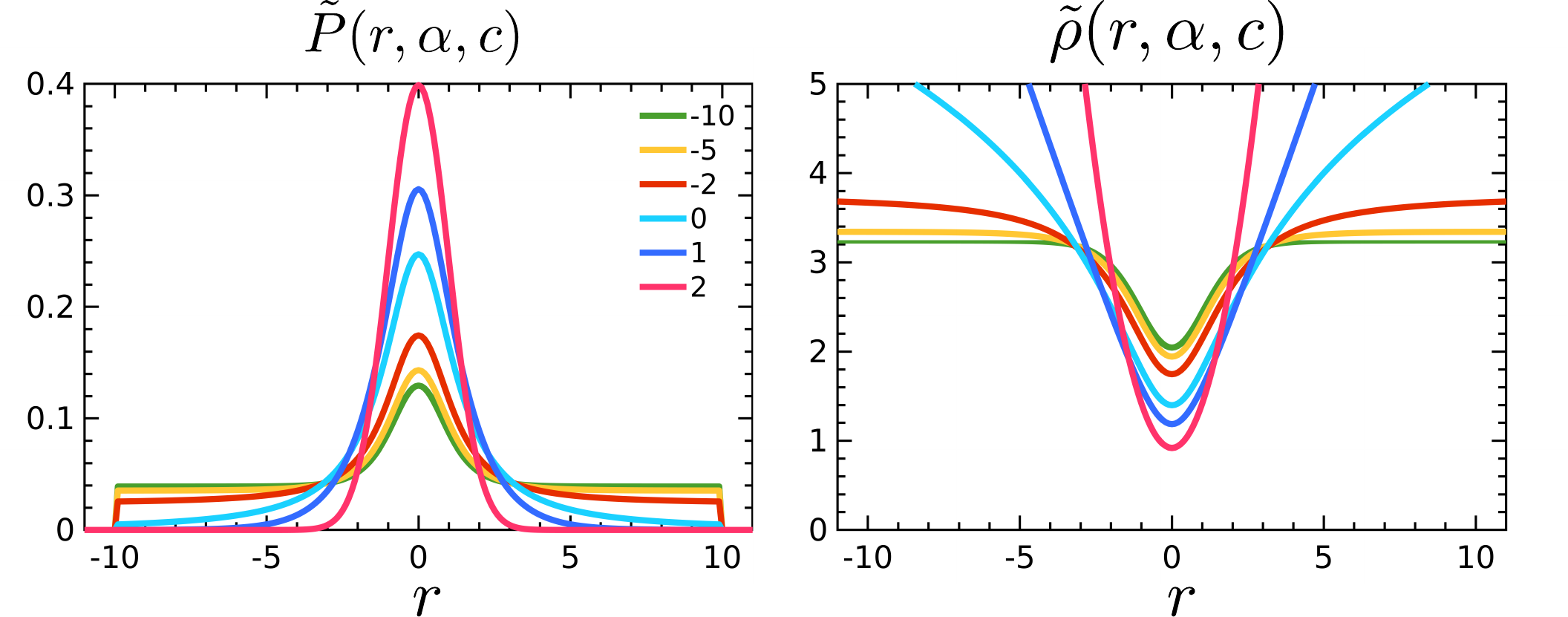}
  \caption{Left: Modified probability distribution $\tilde{\Prob}(r, \alpha, c)$ obtained by truncating $\Prob(r, \alpha, c)$ at $|r| < \tau$. Right: The truncated robust loss $\tilde{\rho}_a(r, \alpha, c)$ allows the automatic tuning of $\alpha$ in its complete range, including $\alpha < 0$.}
  \label{fig:truncated_loss}
\end{figure}

The probability distribution~$\Prob(r,\alpha,c)$ is only defined for $\alpha\geq0$, as the integral in the partition function $Z(\alpha)$ is unbounded for $\alpha<0$. This means that values for $\alpha<0$ cannot be achieved while minimizing the adaptive loss~$\rho_a(\cdot)$ in~\eqref{eq:adaptive_general_loss}. This limits the range of kernels that we can dynamically adapt to. As we can see in~\figref{fig:barron_loss}, the smaller the parameter~$\alpha$ is, the stronger is the down-weighting of outliers. Such a behavior is often desired in situations where a large number of outliers are present in the data.  

In this paper, we propose an extension to the adaptive loss in~\eqref{eq:adaptive_general_loss}, which allows the parameter~$\alpha$ to be dynamically adapted for a larger range of values. We achieve this by limiting the partition to bounded values.
To re-gain the kernels corresponding to the negative range of~$\alpha$ with the adaptive loss function, we compute an approximate partition function~$\tilde{Z}(\alpha)$ as
\begin{eqnarray}
\label{eq:truncated_partition}
\tilde{Z}(\alpha)&=&\int_{-\tau}^{\tau} e^{-\rho(r, \alpha, 1)}dr,
\end{eqnarray} 
where~$\tau$ is the truncation limit for approximating the integral. 
This results in a finite partition~$\tilde{Z}(\alpha)$ for all~$\alpha$ as the integral is computed within the limits $[-\tau, \tau]$. We use this to define our truncated loss function as
\begin{eqnarray}
\label{eq:truncated_general_loss}
\tilde{\rho}_{a}(r,\alpha,c)= \rho(r,\alpha,c) + \text{log}\,c\tilde{Z}(\alpha).
\end{eqnarray}

The truncated probability distribution $\tilde{\Prob}(r, \alpha, c)$ and the corresponding truncated loss $\tilde{\rho}_{a}(r,\alpha,c)$ is shown in~\figref{fig:truncated_loss}.
Since the truncated loss is defined for all values of $\alpha$ including $\alpha<0$, we can adapt $\alpha$ in its entire range during the optimization procedure.  We discuss the effect of the truncation of the loss function below in \secref{sec:errorloss}.

\subsection{Optimization of $\alpha$ via Alternating Minimization}
\label{sec:optimization_alpha}
We propose to solve the joint optimization problem over~$\theta$ and~$\alpha$ defined in~\eqref{eq:joint_robust_loss} in an iterative manner using an alternating minimization procedure. The  procedure alternates between two steps: 
(i) the first step where the maximum likelihood value for $\alpha$ is computed, and
(ii) the second step, where the optimal parameters for the model given the $\alpha$ from the previous step is computed. This can be seen as a variation of a coordinate descent approach.
By solving the joint optimization in this manner, we decouple the estimation of the robust kernel parameter~$\alpha$ from the original optimization problem. This allows to us solve for the model parameters~$\theta$ in the same way as before $\alpha$ was introduced. 

We estimate the parameters~$\alpha$ in the first step by minimizing the negative log-likelihood of observing the current residuals, 
\begin{eqnarray}
\label{eq:l_alpha}
L(\alpha)&=& -\sum_{i=1}^{N} \text{log}\,\Prob(r_i(\theta), \alpha, c) \\
         &=& \sum_{i=1}^{N} \text{log}\, c\tilde{Z}(\alpha) + \rho_a(r_i(\theta), \alpha, c),
\end{eqnarray}
i.e.,
\begin{eqnarray}
\label{eq:mle_alpha}
\alpha^* &=& \argmin_{\alpha} L(\alpha).
\end{eqnarray}

The solution to~\eqref{eq:mle_alpha} can be obtained by setting its first derivative~$\frac{dL(\alpha)}{d\alpha}= 0$. Since its not possible to derive the partition function $\tilde{Z}(\alpha)$ analytically, we settle for a numerical solution. As $\alpha$ is a scalar value,~$L(\alpha)$ can be minimized simply by performing a 1-D grid search for~$\alpha \in [\alpha_\mathit{min}, 2]$. 

\begin{algorithm}[t]
  {
  \vspace{0.5cm}
  \caption{Optimization with adaptive robust kernel}
  \label{alg:scheme}
  \begin{algorithmic}[1]
  \State Initialize $\theta^0=\theta, \alpha^0=2, c$ 
  \While{!converged} 
      \State Step 1: Minimize for $\alpha$
      \State $\alpha^t = \argmin_{\alpha} - \sum_{i=1}^{N} \text{log}\,\Prob(r_i(\theta^{t-1}), \alpha^{t-1}, c)$
      \State Step 2: Minimize robust loss using IRLS 
      \State $\theta^t = \argmin_{\theta} \sum_{i=1}^{N} \rho({r_i(\theta), \alpha^t, c}), $
  \EndWhile
  \end{algorithmic}
  }
  \end{algorithm}

In terms of a practical implementation, we chose lower bound~$\alpha_\mathit{min}=-10$ as its difference to the corresponding weights for~$\alpha=-\infty$ for large residuals~$(|r|>\tau)$ is small. The maximum value for $\alpha$ is set to $2$ as this corresponds to the standard least squares problem. The scale $c$ of the robust loss is fixed beforehand and not adapted during the optimization. This value for~$c$ is usually fixed based on the measurement noise for an inlier observation~$z$ and the accuracy of the initial solution. To be computationally efficient, we  pre-compute~$\tilde{Z}(\alpha)$ as a lookup table for values $\alpha \in [\alpha_\mathit{min},2]$ with a resolution of $0.1$ and use the lookup table during optimization.  
This leads us to the overall minimization approach shown in \algref{alg:scheme}.

\subsection{Effect of Using the Truncated Loss}
\label{sec:errorloss}

The second step, which is the \emph{minimization} in the NLS estimation, is not affected by the truncated loss approximation. It can, however, affect the  first step, \ie,  determining the parameter~$\alpha$. By using our truncated loss, we are implicitly assuming that no outliers have a residual $|r|>\tau$ during the first step. If we choose a large enough value for $\tau$, the error that we introduce affects situations with large outliers only and therefore results in small  values of~$\alpha$. The effect of small $\alpha$ values such as $\alpha=-10$ vs.~$\alpha=-\infty$ on the optimization, however, is negligible as the outliers will be down-weighted to basically zero. 
We observe in our experiments that by setting choosing $\tau=10c$, we are able to deal with almost all of the outlier distributions occurring in practice for ICP, SLAM, BA applications.

\setlength\tabcolsep{4.75pt}\begin{table*}[ht]
  \centering
  \caption{Results on KITTI odometry datasets [Relative rot. error in $\mathrm{degrees}$ per $100$\,m / relative trans. error in $\%$]}
  \scriptsize{
  \begin{tabular}{lcccccccccccc}
  \toprule
  & \multicolumn{11}{c}{Sequence} &  \\
  Approach & 00 & 01 & 02 & 03 & 04 & 05 & 06 & 07 & 08 & 09 & 10 & Average \\
  \midrule
  Our Approach & $1.5$/$2.8$ & $1.3$/$\mathbf{3.8}$ & $0.91$/$\mathbf{1.8}$ & $1.5$/$1.9$ & $0.81$/$\mathbf{0.95}$ & $0.97$/$1.7$ & $0.51$/$1.1$ & $2.1$/$2.6$ & $1.3$/$2.7$ & $0.80$/$\mathbf{1.4}$ & $1.3$/$\mathbf{1.7}$  & $1.18$/$\mathbf{2.03}$ \\
  \midrule
  Adaptive Kernel & $1.6$/$3.0$ & $1.2$/$6.7$ & $0.93$/$1.9$ & $1.4$/$1.8$ & $0.82$/$1.0$ & $0.97$/$1.8$ & $0.51$/$1.1$ & $2.2$/$2.7$ & $1.3$/$2.8$ & $0.88$/$\mathbf{1.4}$ & $1.2$/$\mathbf{1.7}$  & $1.19$/$2.35$ \\
  \multicolumn{12}{p{0.9\linewidth}}{(Barron~\cite{barron2019cvpr})} \\
  \midrule
  Fixed Kernel & $0.93$/$\mathbf{2.1}$ & $1.2$/$4.5$ & $0.79$/$2.3$ & $0.7$/$\mathbf{1.4}$ & $1.1$/$49$ & $0.79$/$1.5$ & $0.64$/$\mathbf{0.95}$ & $1.2$/$\mathbf{1.8}$ & $0.96$/$\mathbf{2.5}$ & $0.78$/$1.9$ & $0.97$/$1.8$  & $0.92$/$6.34$ \\
  \multicolumn{12}{p{0.9\linewidth}}{(Huber)} \\
  \midrule
  Fixed Kernel & $1.8$/$3.4$ & $1.3$/$\mathbf{3.8}$ & $1.0$/$1.9$ & $1.5$/$2.0$ & $0.88$/$1.2$ & $0.98$/$1.7$ & $0.62$/$1.3$ & $2.6$/$3.0$ & $1.5$/$3.0$ & $1.0$/$1.6$ & $1.3$/$1.9$  & $1.32$/$2.27$ \\
  \multicolumn{12}{p{0.9\linewidth}}{(Geman-McClure)} \\
  \midrule
  Hand-Crafted & $0.9$/$\mathbf{2.1}$ & $1.2$/$4.0$ & $0.8$/$2.3$ & $0.7$/$\mathbf{1.4}$ & $1.1$/$11.9$ & $0.8$/$1.5$ & $0.6$/$1.0$ & $1.2$/$\mathbf{1.8}$ & $1.0$/$\mathbf{2.5}$ & $0.8$/$1.9$ & $1.0$/$1.8$  & $0.9$/$2.90$ \\
  \multicolumn{12}{p{0.9\linewidth}}{Outlier Rejection~\cite{behley2018rss}} \\
  \bottomrule
  \end{tabular}
  }
  \label{tab:kitti-eval}
  \end{table*}

\section{Experimental Evaluation}
\label{sec:exp}

In the experimental section, we evaluate the performance of our adaptive robust kernel approach on two common problems in robotics, (i) registration using Iterative Closest Point (ICP) algorithm, and (ii) bundle adjustment as an example state estimation task. The experiments are designed to evaluate effectiveness of our approach in presence of strong outliers, and showcase its applicability for common NLS problems. We compare the performance of our approach against a hand-crafted outlier rejection mechanism using fixed robust kernels and the original adaptive kernel formulation by Barron~\cite{barron2019cvpr} on a benchmarking dataset and conduct analysis on simulated data to understand the convergence properties.  

\subsection{Application to Iterative Closest Point}

The first experiment is designed to show the advantages of our approach for LiDAR-based registration in form of ICP. We integrated our truncated adaptive robust kernel into an existing SLAM system, called surfel-based mapping (SuMa)~\cite{behley2018rss}, which performs point-to-plane projective ICP for 3D LiDAR scans. The ICP registration is performed in a frame-to-frame fashion on consecutive scans. We compare the performance of our approach against two fixed robust kernels, \ie, Huber and Geman-McClure, as well as to a hand-crafted outlier rejection scheme as used in the original implementation of SuMa~\cite{behley2018rss}. This hand-crafted scheme combines a Huber kernel with an additional outlier rejection step that removes all correspondences, which have a distance of more than $2\,m$ or which have an angular difference greater than $30^\circ$ between the estimated normals of observations and the corresponding normals of the surfels. Finally, we also compare it to the original adaptive kernel by Barron~\cite{barron2019cvpr}.  

We evaluate all these approaches on the odometry datasets of the KITTI vision benchmark~\cite{geiger2012cvpr} and summarize the results in~\tabref{tab:kitti-eval}. The best performance in terms of relative translation error for each sequence is highlighted in bold. We observe that our proposed approach, which does \emph{not} require an outlier rejection step at all, performs better or is on-par with fixed kernel plus outlier rejection scheme for many of the sequences. At the same time, using only the fixed kernel without the outlier rejection step fails for some of the sequences. In particular for Sequence 04, both the fixed kernel using Huber and the hand-crafted outlier rejection scheme fail, whereas our approach performs the best on this sequence. Our approach performs slightly better with respect to the adaptive kernel by Barron~\cite{barron2019cvpr}, with the biggest gain in Sequence 01, which requires negative $\alpha$ values to deal with the outliers coming from dynamic objects in the scene. On average over all the sequences, our approach provides the best accuracy in terms relative translation error. Here, we note that our approach is not the best in terms of relative rotational error, but is around the 1\%~mark, which is on-par with other approaches.  
These results are promising as by using our adaptive robust kernel, we do not need any hand-crafted outlier rejection mechanism, which in practice requires manual tuning for new data or different sensor configurations.  

\begin{figure*}[ht]
  \centering
  \includegraphics[width=\linewidth]{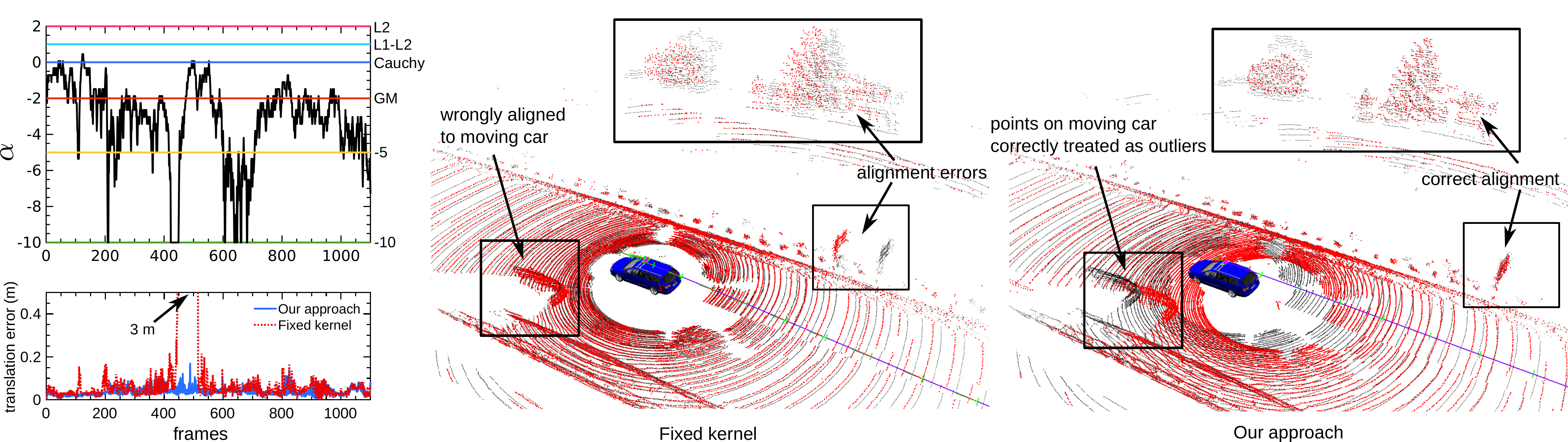}
  \caption{Registration by ICP and our approach on a challenging dataset for ICP. Top-left: Plot showing $\alpha$ values estimated at each frame based on the residual distribution for KITTI 01 sequence. Bottom-Left: Translation error~(in meters) for our approach and fixed kernel ICP. We use Huber kernel as used as the fixed robust kernel in this example.
  We observe lower $\alpha$ values~(stronger outlier down-weighting) for scan matches with outliers arising from dynamic objects in the scene. Middle-right: ICP result from an example frame where ICP converges for adaptive kernel whereas it diverges for a fixed kernel.}
  \label{fig:exp_icp_kitti}
\end{figure*}

As a qualitative evaluation, we illustrate the advantage of using the adaptive robust kernel for a challenging dataset~(Sequence 01), which contains several moving cars moving with the vehicle itself along the highway with little additional geometric structures. In~\figref{fig:exp_icp_kitti}~(top-left), we plot the values of $\alpha$ for each iteration while mapping  the sequence. We observe that $\alpha$ adapts to smaller and more negative values whenever there are more outliers, which arise mainly from moving vehicles in the scan. This effect can be seen in~\figref{fig:exp_icp_kitti}~(bottom-left) where the translation error for the fixed Huber kernel increases as it cannot handle the outlier situation well. At the same time, the error remains small for our adaptive kernel. 

The two 3D scenes show the registrations at the same point in time, once computed with the adaptive kernel and once with a fixed one. The adaptive kernel results in a successful alignment while the fixed kernel fails to find the correct solution due to the outlier in the data association, see~\figref{fig:exp_icp_kitti}~(middle and right). The adaptive kernel-based ICP can correctly treat the observations belonging to the moving car as outliers, and nullify their effect during optimization automatically. For this sequence, we note that for large portions of the scans, $\alpha$ is negative and even reaches down to $\alpha_{\text{min}}=-10$ in some instances. This suggests that our truncated adaptive loss proposed in~\eqref{eq:truncated_general_loss} is critical for the successful application of ICP as it enables using values $\alpha<0$, whereas the original formulation of the adaptive loss is limited to $\alpha \in [0,2]$.  Thus, our approach greatly supports ICP-based registration as it avoids hand-crafted outlier strategies and adapts to the outlier challenges present in each pairs of scans automatically.

\subsection{Application to Bundle Adjustment (BA)}
\label{ssec:app_ba}

The second experiment is designed to illustrate the performance of our approach and its advantages for the bundle adjustment problem using a monocular camera. We integrated the adaptive robust kernel to an existing bundle adjustment framework proposed by Schneider \etal~\cite{schneider2012isprs}. The initial estimate for camera poses and 3D points is obtained by three commonly used steps. First, extract SIFT features and compute possible matches between all image pairs. Second, compute the relative orientation using Nister's 5-point algorithm~\cite{nister2004pami} together with RANSAC for outlier rejection and chaining the subsequent images to obtain the initial camera trajectory. Third, compute the 3D points as described by L\"abe \etal~\cite{laebe2006tggd} given the camera trajectory of the second step.

\begin{figure}[t]
  \centering
  \includegraphics[width=0.49\columnwidth]{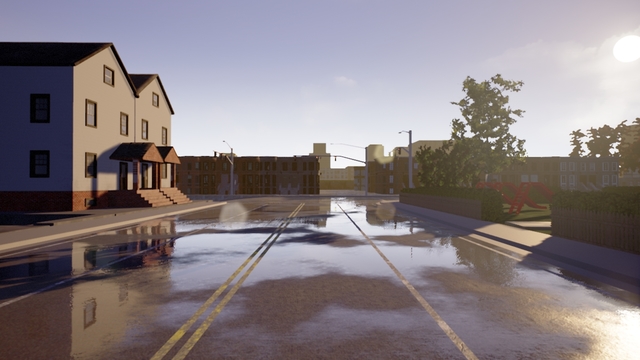}
  \includegraphics[width=0.49\columnwidth]{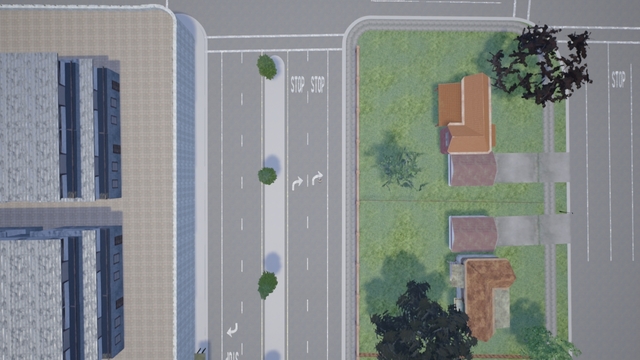} \\
  \vspace{1mm}
  \includegraphics[width=0.49\columnwidth]{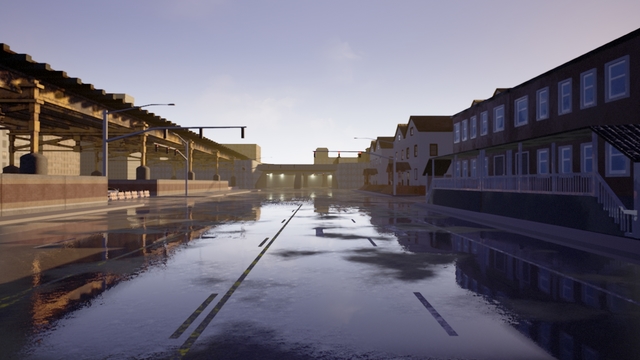}
  \includegraphics[width=0.49\columnwidth]{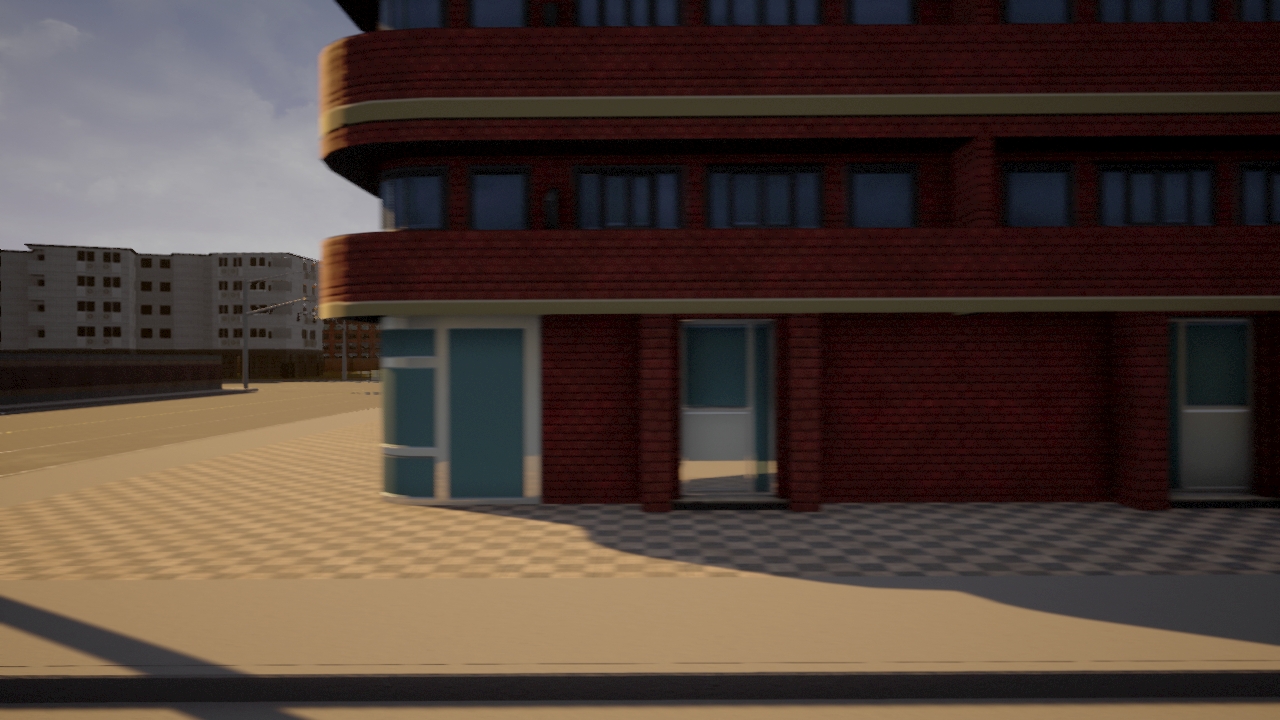}
  \caption{Examples images from CARLA simulated datasets.
  Top left: front-looking image mounted on a car from the first dataset.
  Top right: UAV image in a nadir view looking downwards from the second dataset. 
  Bottom left: front-looking image with strong shadows and reflections from the third dataset.
  Bottom right: side facing image from the fourth dataset where portions of the image have significant motion blur.}
  \label{fig:carla_datasets}
  \vspace{0.1cm}
\end{figure}

To test the bundle adjustment performance, we created four datasets covering different scenarios using the CARLA simulator~\cite{dosovitskiy2017rl} generating near-realistic images. The advantage of the simulator is that ground truth information for the cameras poses is available. The first dataset contains images from a front looking camera mounted on a car, the second dataset simulates downward looking aerial images from a UAV, the third dataset contains images where around half of each image shows strong shadows, and the fourth dataset simulates side-ward looking camera where close-by objects suffer from significant motion blur. We have generated these datasets to cover situations where feature matching is challenging, thereby resulting in a large number of outliers. Example images from the all the four datasets are depicted in \figref{fig:carla_datasets}.

For each of these datasets, we evaluate the bundle adjustment results by comparing the performance of our approach against squared error loss as well as the standard Huber loss as a fixed kernel. 
We also compare the results to two adaptive kernels which include the original adaptive kernel by Barron~\cite{barron2019cvpr} as well as a prior work by Agamennoni~\etal~\cite{agamennoni2015icra} based on a family of robust kernels derived from an elliptical distribution. We re-implemented Agamennoni's~\cite{agamennoni2015icra} approach using the parameters provided for the visual SLAM problem in the paper as it is the closest one to the bundle adjustment problem evaluated here.    
To evaluate the performance, we compute the accuracy of the camera poses estimated by each of the approached by comparing them against the ground truth poses from the simulator as described by Dickscheid \etal~\cite{dickscheid2008isprs}. 
This difference is computed by estimating the optimal transformation between the bundle adjustment result and the ground truth using all 6 DoF pose parameters with the approach by Dickscheid \etal~\cite{dickscheid2008isprs}. Note that in this analysis, we perform the ground truth comparison only based on the camera poses and do not consider the 3D point as they have been extracted using the SIFT descriptor from the simulated images and thus no ground truth 3D information is available.

\begin{figure}
  \centering
  \includegraphics[width=\columnwidth]{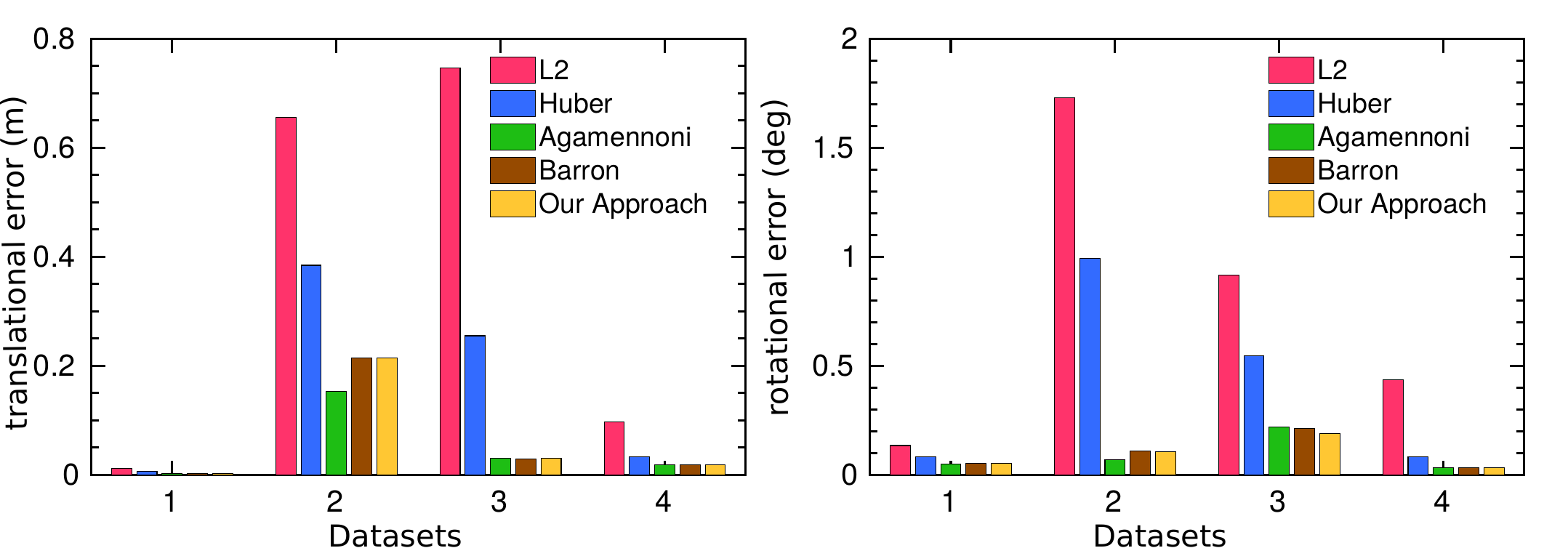}
  \caption{Performance of different fixed and adaptive robust kernels for bundle adjustment problem on CARLA datasets. We show the translation error (in meters) on the left and rotational errors (in degrees) on the right.}
  \label{fig:ba_accuracy}
\end{figure}

We show the results of the bundle adjustment experiment for all the four datasets in ~\figref{fig:ba_accuracy}. Here, we see that our approach has a lower translation and rotational error than using squared error or the fixed Huber kernel. We obtain a translation and rotational error, which is between $2$ to $5$ times better as compared to using Huber depending on the dataset. 
Our approach also shows a similar performance as compared to the adaptive kernel by Barron~\cite{barron2019cvpr} with a marginal gain on the rotational error some datasets. We believe that the relatively similar performance of the two approaches in these experiments is due to the fact that large negative $\alpha$'s were not needed as often to deal with the outlier distributions in our example datasets.
We also obtained similar accuracy results as compared to Agamennoni's approach for most of the datasets, except for second dataset where it outperforms our approach in terms of the translational error. While showing better accuracy results on these datasets, we note that Agamennoni's approach comes at computational overhead. This approach has a two step process to determine the best robust kernel given a residual distribution. It first computes the best hyper-parameter value (similar to $\alpha$ in this paper) for each of the kernels which can be derived from an elliptical distribution (five such kernels are provided in the paper), and then a second step to performed by computing the Kullback-Leibler divergence to determine the best kernel. 

\begin{figure}
  \centering
   \subfigure[Squared Loss]{\includegraphics[trim=0 42 0 42,clip,width=0.49\columnwidth]{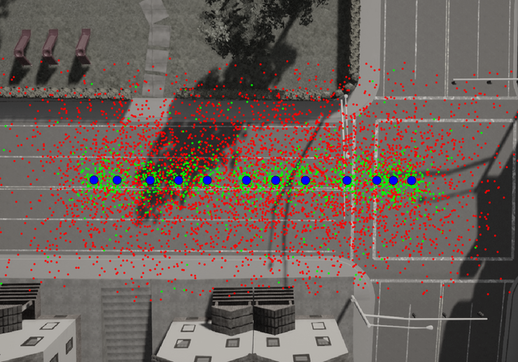}}
   \subfigure[Huber]{\includegraphics[trim=0 42 0 42,clip,width=0.49\columnwidth]{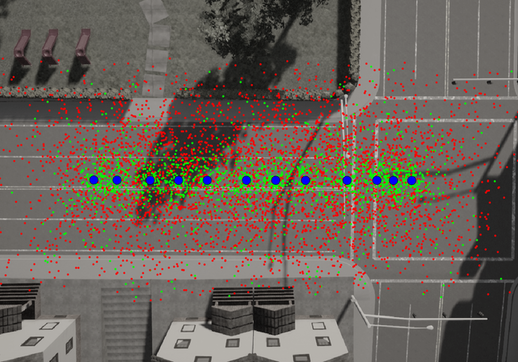}}
   \subfigure[Geman-McClure]{\includegraphics[trim=0 42 0 42,clip,width=0.49\columnwidth]{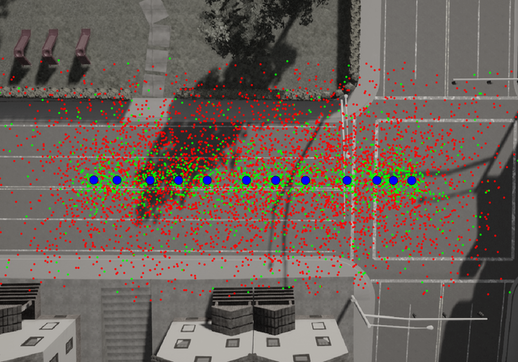}}
   \subfigure[Our Approach]{\includegraphics[trim=0 42 0 42,clip,width=0.49\columnwidth]{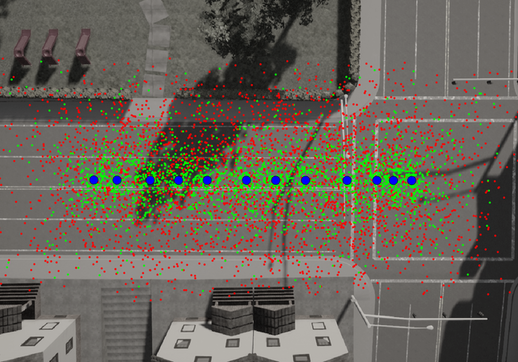}}
   \caption{Convergence analysis for BA. Green points indicate poses for which BA converged, whereas red points indicate divergence. The blue circles represent the ground truth camera poses.}
  \label{fig:ba_convergence}
\end{figure}

The last experiment in this section is designed to  analyze the influence of our approach on the convergence properties of BA. A large basin of convergence is important for robust operation, especially for BA due to the missing range information with the image data. We initialized the bundle adjustment procedure by adding significant noise to the initial camera poses, \ie, $\sigma\in[0.1\,\text{m}, 5\,\text{m}]$ to the ground truth poses of the camera. The noise in the camera poses is propagated to the 3D points during the forward intersection step. We sample 20 instances of each noise level~(500 instances in total) and run the bundle adjustment for our approach, using squared loss, the Geman-McClure as well as the Huber kernel. We consider the adjustment to have converged if the final RMS error of the camera center is less than $1\,$cm from the true position. We visualize the results in~\figref{fig:ba_convergence} where the poses from which the BA has converged are shown in green and the ones that caused divergence in red. We can clearly see that our approach has a larger convergence radius as the green points are spread over a larger area  compared to the squared loss or fixed Huber or Geman-McClure kernel. We obtain successful convergence rate for $45\%$ of all instances for our approach against $24.8\%$ for squared loss, $33\%$ for Huber, and $28.2\%$ for Geman-McClure. Overall, the experiments suggest that by using our approach, we can obtain a more accurate estimate and have a larger convergence area as compared to a fixed kernel. Thus, our approach is an effective and useful approach for optimization in bundle adjustment problems.

\subsection{Effect of the Truncation Parameter}

\begin{figure}
  \centering
  \includegraphics[width=\columnwidth]{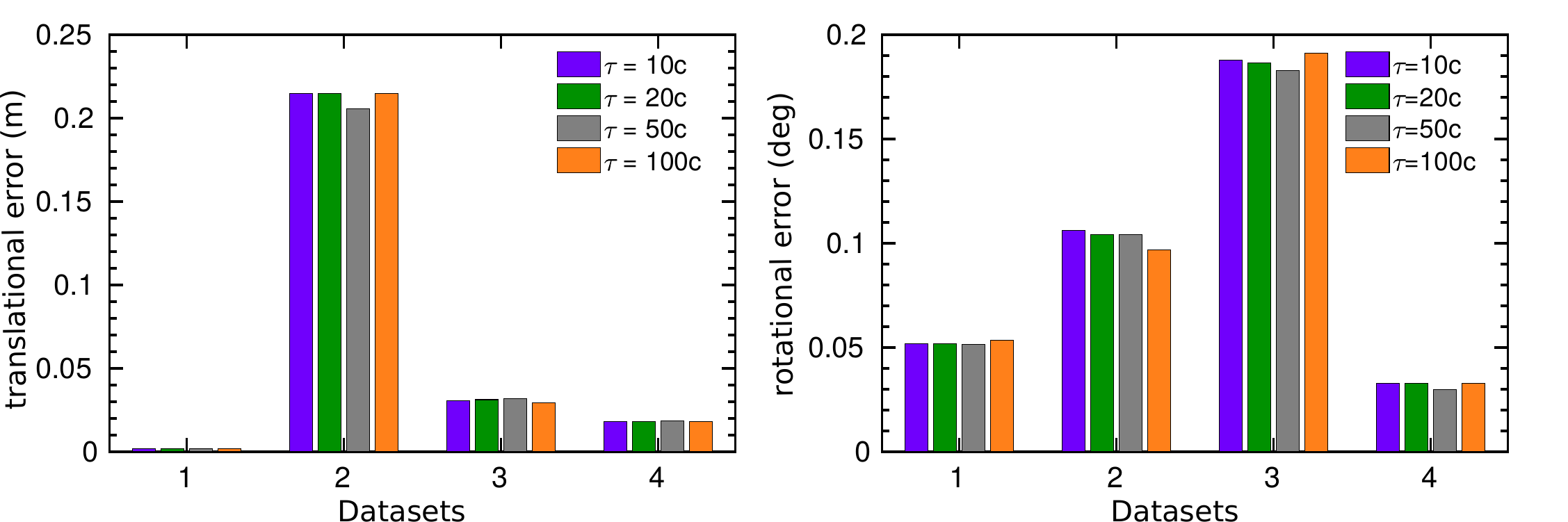}
  \caption{Effect of the truncation parameter $\tau$ on the accuracy on the bundle adjustment task.}
  \label{fig:ba_accuracy_tau}
 \end{figure}

In this experiment, our goal is to analyze the effect of the truncation parameter $\tau$ on the state estimation task. The parameter $\tau$ is used to define the integral limits for the partition function in~\eqref{eq:truncated_partition}. By choosing a value of $\tau$ from the set $\{10c, 20c, 50c, 100c\}$, we define a truncated partition function $\tilde{Z}(.)$ for each value of $\tau$.
This results in multiple robust kernels $\tilde{\rho}_{a}(.)$ which are then used for solving the bundle adjustment problem on the four datasets as described in~\secref{ssec:app_ba}.
The results for these experiments are shown in \figref{fig:ba_accuracy_tau}. We observe that the accuracy results using different values of $\tau$ is similar for all the datasets. The maximum difference in the translation error is about 5\%, and the difference in rotational error is 8\% using different values for $\tau$. 
We also note that during the experiments, the $\alpha$ estimated in Step 1 of~\algref{alg:scheme} of the optimization process belongs to a similar range of $\alpha$ values for each of the $\tau$ used. These results in this experiment suggests that the approach is not critically sensitive to the value of the truncation parameter used.

\section{Limitations and Future Work}
\label{sec:limitations}
In this paper, we have mainly focused on extending the generalized robust kernel formulation by Barron~\cite{barron2019cvpr} for its use in common state estimation problems in robotics. There are several interesting directions in which this work can be extended. We see two such aspects, which offer space for further investigations:

\textit{Adapting scale parameter $c$}: In our current implementation, we use a fixed scale parameter $c$, which is set based on the measurement noise for an inlier observation. In principle, both, the shape parameter~$\alpha$ and the scale parameter~$c$ can be adapted simultaneously. In practice, however, learning both these parameters jointly becomes tricky as each of them can individually find \emph{some explanation} of the residual distribution. Typically, a large change in the residuals can be captured by a rather small change in $c$ without affecting~$\alpha$ at all. This results in a situation where the shape parameter~$\alpha$ cannot be estimated properly. This suggests that a different scheme needs to be developed to adapt the parameter~$\alpha$ and the scale parameter $c$ jointly to respond to a change in the outlier distribution.
  
\textit{Use of multiple $\alpha$ parameters}: We use a single~$\alpha$ value to capture the outlier distribution for the overall optimization problem in our example applications. For example, in the ICP case, we estimate only one~$\alpha$ value for all the points for a scan pair. This means, the $\alpha$ value is adapted over time (between different scan pairs), however, is the same for all the points in the scan pair at one timestep~$t$.  By estimating a different~$\alpha$ value for groups of points belonging to different objects (e.g. vegetation, road, vehicles etc.) in the scan, we can model an inlier/outlier weight for each group of points in addition to time. Similarly, in the bundle adjustment application, we adapt $\alpha$ at each iteration of the optimization procedure. Here, one could estimate different $\alpha$ values for each sub-block of the optimization problem, where each of the sub-block could consist of the camera poses and measurements of a particular location in the environment.

\textit{Use of an alternative regularization term for the truncated loss}: 
The truncated partition function $\tilde{Z}(\alpha)$ can be seen as playing the role of a regularizer for~$\alpha$ in~\eqref{eq:truncated_general_loss}. An alternative approach to regularizing $\alpha$ would be to replace the truncated integral term $\tilde{Z}(\alpha)$  with a suitable regularizer that is defined for any $\alpha$ in the range $[-\infty, 2]$. 
This is possible as there is no strict requirement that a robust loss $\rho(\cdot)$ must correspond to the negative log-likelihood of a probability distribution function as we have defined in this paper. This opens up the possibility of designing regularization terms that have potentially better outlier rejection properties, and provides an interesting direction for future work.

\section{Conclusion}
\label{sec:conclusion}

In this paper, we presented a novel approach to robust optimization that
avoids the need to commit to a fixed robust kernel and potentially has a broad
application area for state estimation in robotics. We proposed the use of a
generalized robust kernel that can adapt its shape with an additional
parameter that has recently been proposed by  Barron~\cite{barron2019cvpr}. We
modified the original formulation, which enables us to use the adaptive kernel
also in situations with strong outliers. We integrated our adaptive kernel
into and tested it for two popular state estimation problems in robotics,
namely ICP and bundle adjustment. The experiments showcase that we are better
or on-par with fixed kernels such as Huber or Geman-McClure but do not require
hand-crafted outlier rejection schemes in the case of ICP and can increase the
radius of convergence for bundle adjustment. We believe that several other
problems in robotics, which rely on robust least-squares estimation, can
benefit from our proposed approach.

\bibliographystyle{plain}

\begin{thebibliography}{10}

\bibitem{agamennoni2015icra}
G.~Agamennoni, P.T. Furgale, and R.~Siegwart.
\newblock {Self-Tuning M-Estimators}.
\newblock In {\em Proc.~of the IEEE Intl.~Conf.~on Robotics \& Automation
  (ICRA)}, 2015.

\bibitem{agarwal2014ram}
P.~Agarwal, W.~Burgard, and C.~Stachniss.
\newblock {A Survey of Geodetic Approaches to Mapping and the Relationship to
  Graph-Based SLAM}.
\newblock {\em IEEE Robotics and Automation Magazine (RAM)}, 2014.

\bibitem{agarwal2013icra}
P.~Agarwal, G.D. Tipaldi, L.~Spinello, C.~Stachniss, and W.~Burgard.
\newblock {Robust Map Optimization using Dynamic Covariance Scaling}.
\newblock In {\em Proc.~of the IEEE Intl.~Conf.~on Robotics \& Automation
  (ICRA)}, 2013.

\bibitem{agarwal2010ceres}
S.~Agarwal, K.~Mierle, and Others.
\newblock {Ceres Solver}.
\newblock http://ceres-solver.org, 2010.

\bibitem{babin2019icra}
P.~Babin, P.~Giguere, and F.~Pomerleau.
\newblock Analysis of robust functions for registration algorithms.
\newblock In {\em Proc.~of the IEEE Intl.~Conf.~on Robotics \& Automation
  (ICRA)}, 2019.

\bibitem{barron2019cvpr}
J.~T. Barron.
\newblock {A General and Adaptive Robust Loss Function}.
\newblock In {\em Proc.~of the IEEE Conf.~on Computer Vision and Pattern
  Recognition (CVPR)}, 2019.

\bibitem{behley2018rss}
J.~Behley and C.~Stachniss.
\newblock {Efficient Surfel-Based SLAM using 3D Laser Range Data in Urban
  Environments}.
\newblock In {\em Proc.~of Robotics: Science and Systems (RSS)}, 2018.

\bibitem{black1996ijcv}
M.~J. Black and A.~Rangarajan.
\newblock On the unification of line processes, outlier rejection, and robust
  statistics with applications in early vision.
\newblock {\em Intl.~Journal~of Computer Vision (IJCV)}, 19(1):57--91, 1996.

\bibitem{bosse2016fntr}
M.~Bosse, G.~Agamennoni, and I.~Gilitschenski.
\newblock Robust estimation and applications in robotics.
\newblock {\em Foundations and Trends in Robotics}, 4(4):225--269, 2016.

\bibitem{dickscheid2008isprs}
T.~Dickscheid, T.~L\"abe, and W.~F\"orstner.
\newblock Benchmarking automatic bundle adjustment results.
\newblock In {\em Cong.~of the Intl.~Society for Photogrammetry and Remote
  Sensing (ISPRS)}, 2008.

\bibitem{dosovitskiy2017rl}
A.~Dosovitskiy, G.~Ros, F.~Codevilla, A.~Lopez, and V.~Koltun.
\newblock {CARLA}: {An} open urban driving simulator.
\newblock In {\em Proc.~of the Conf.~on Robot Learning}, 2017.

\bibitem{geiger2012cvpr}
A.~Geiger, P.~Lenz, and R.~Urtasun.
\newblock {Are we ready for Autonomous Driving? The KITTI Vision Benchmark
  Suite}.
\newblock In {\em Proc.~of the IEEE Conf.~on Computer Vision and Pattern
  Recognition (CVPR)}, 2012.

\bibitem{huber1964kernel}
P.~J. Huber.
\newblock Robust estimation of a location parameter.
\newblock {\em Annals of Mathematical Statistics}, 35(1):73--101, 1964.

\bibitem{kaess2008tro}
M.~Kaess, A.~Ranganathan, and F.~Dellaert.
\newblock {iSAM}: Incremental smoothing and mapping.
\newblock {\em IEEE Trans. on Robotics (TRO)}, 24(6), 2008.

\bibitem{kummerle2011icra}
R.~K{\"u}mmerle, G.~Grisetti, H.~Strasdat, K.~Konolige, and W.~Burgard.
\newblock {g2o: A general framework for graph optimization}.
\newblock In {\em Proc.~of the IEEE Intl.~Conf.~on Robotics \& Automation
  (ICRA)}, pages 3607--3613, 2011.

\bibitem{laebe2006tggd}
T.~L\"abe and W.~F\"orstner.
\newblock Automatic relative orientation of images.
\newblock In {\em Proc.~of the Turkish-German Joint Geodetic Days}, 2006.

\bibitem{lajoie2019ral}
P.~Lajoie, S.~Hu, G.~Beltrame, and L.~Carlone.
\newblock Modeling perceptual aliasing in {SLAM} via discrete-continuous
  graphical models.
\newblock {\em IEEE Robotics and Automation Letters (RA-L)}, 2019.

\bibitem{latif2012rss}
Y.~Latif, C.~Cadena, and J.~Neira.
\newblock Robust loop closing over time.
\newblock {\em Proc.~of Robotics: Science and Systems (RSS)}, 2012.

\bibitem{mactavish2015crv}
K.~MacTavish and T.~D. Barfoot.
\newblock At all costs: A comparison of robust cost functions for camera
  correspondence outliers.
\newblock In {\em Proc.~of the Conf.~on Computer and Robot Vision}, pages
  62--69, 2015.

\bibitem{nister2004pami}
D.~Nist{\'e}r.
\newblock An efficient solution to the five-point relative pose problem.
\newblock {\em IEEE Trans.~on Pattern Analalysis and Machine Intelligence
  (TPAMI)}, 26(6):756--770, 2004.

\bibitem{schneider2012isprs}
J.~Schneider, F.~Schindler, T.~L\"abe, and W.~F\"orstner.
\newblock Bundle adjustment for multi-camera systems with points at infinity.
\newblock In {\em ISPRS Annals of Photogrammetry, Rem. Sens. and Spatial Inf.
  Sci.}, volume I-3, 2012.

\bibitem{stachniss2016handbook-slamchapter}
C.~Stachniss, J.~Leonard, and S.~Thrun.
\newblock {\em {Springer Handbook of Robotics, 2nd edition}}, chapter
  Chapt.~46: Simultaneous Localization and Mapping.
\newblock Springer Verlag, 2016.

\bibitem{sunderhauf2012iros}
N.~S{\"u}nderhauf and P.~Protzel.
\newblock Switchable constraints for robust pose graph slam.
\newblock In {\em Proc.~of the IEEE/RSJ Intl.~Conf.~on Intelligent Robots and
  Systems (IROS)}, pages 1879--1884, 2012.

\bibitem{yang2020ral}
H.~Yang, P.~Antonante, V.~Tzoumas, and L.~Carlone.
\newblock Graduated non-convexity for robust spatial perception: From
  non-minimal solvers to global outlier rejection.
\newblock {\em IEEE Robotics and Automation Letters (RA-L)}, 5(2):1127--1134,
  2020.

\bibitem{yang2020arxiv}
H.~Yang, J.~Shi, and L.~Carlone.
\newblock {TEASER: Fast and Certifiable Point Cloud Registration}.
\newblock {\em arXiv preprint arXiv: 2001.07715}, 2020.

\bibitem{zach2014eccv}
C.~Zach.
\newblock Robust bundle adjustment revisited.
\newblock In {\em Proc.~of the Europ.~Conf.~on Computer Vision (ECCV)}, pages
  772--787, 2014.

\bibitem{zhang1997ivc}
Z.~Zhang.
\newblock Parameter estimation techniques: A tutorial with application to conic
  fitting.
\newblock {\em Image and Vision Computing}, 15:59--76, 1997.

\end{thebibliography}

\end{document}